\documentclass[sigconf]{acmart}
\usepackage{balance}
\def\BibTeX{{\rm B\kern-.05em{\sc i\kern-.025em b}\kern-.08emT\kern-.1667em\lower.7ex\hbox{E}\kern-.125emX}}
    
\usepackage{booktabs} % For formal tables
\usepackage{hyperref}
\usepackage{url}
\usepackage{graphicx}
\usepackage{amsmath}
\usepackage{amssymb}
\usepackage{multirow}
\usepackage{array}
\usepackage{blindtext}
\usepackage{multicol}
\usepackage{tabularx}
\usepackage{amsmath}
\usepackage{svg}
\usepackage{caption} 
\usepackage{pgfplots}
\usepackage[title]{appendix}
\usepackage{arydshln}

\let\vec\mathbf
\DeclareMathOperator*{\argmin}{arg\,min}

\copyrightyear{2019}
\acmYear{2019}
\setcopyright{acmcopyright}
\acmConference[ICMR '19]{International Conference on Multimedia Retrieval}{June 10--13, 2019}{Ottawa, ON, Canada}
\acmBooktitle{International Conference on Multimedia Retrieval (ICMR '19), June 10--13, 2019, Ottawa, ON, Canada}
\acmPrice{15.00}
\acmDOI{10.1145/3323873.3325028}
\acmISBN{978-1-4503-6765-3/19/06}

\begin{document}

% The "title" command has an optional parameter, allowing the author to define a "short title" to be used in page headers.
\title{Context-Aware Embeddings for Automatic Art Analysis}

% The "author" command and its associated commands are used to define the authors and their affiliations.
% Of note is the shared affiliation of the first two authors, and the "authornote" and "authornotemark" commands
% used to denote shared contribution to the research.
\author{Noa Garcia}
\affiliation{%
  \institution{Institute for Datability Science}
  \city{Osaka University}
  \country{Japan}}
\email{noagarcia@ids.osaka-u.ac.jp}

\author{Benjamin Renoust}
\affiliation{%
  \institution{Institute for Datability Science}
  \city{Osaka University}
  \country{Japan}}
\email{renoust@ids.osaka-u.ac.jp}

\author{Yuta Nakashima}
\affiliation{%
  \institution{Institute for Datability Science}
  \city{Osaka University}
  \country{Japan}}
\email{n-yuta@ids.osaka-u.ac.jp}

%
% By default, the full list of authors will be used in the page headers. Often, this list is too long, and will overlap
% other information printed in the page headers. This command allows the author to define a more concise list
% of authors' names for this purpose.
\renewcommand{\shortauthors}{Garcia, et al.}

%
% The abstract is a short summary of the work to be presented in the article.
\begin{abstract}
Automatic art analysis aims to classify and retrieve artistic representations from a collection of images by using computer vision and machine learning techniques. In this work, we propose to enhance visual representations from neural networks with contextual artistic information. Whereas visual representations are able to capture information about the content and the style of an artwork, our proposed context-aware embeddings additionally encode relationships between different artistic attributes, such as author, school, or historical period. We design two different approaches for using context in automatic art analysis. In the first one, contextual data is obtained through a multi-task learning model, in which several attributes are trained together to find visual relationships between elements. In the second approach, context is obtained through an art-specific knowledge graph, which encodes relationships between artistic attributes. An exhaustive evaluation of both of our models in several art analysis problems, such as author identification, type classification, or cross-modal retrieval, show that performance is improved by up to 7.3\% in art classification and 37.24\% in retrieval when context-aware embeddings are used.
\end{abstract}



%
% Keywords. The author(s) should pick words that accurately describe the work being
% presented. Separate the keywords with commas.

%
% This command processes the author and affiliation and title information and builds
% the first part of the formatted document.
\maketitle

\section{Introduction}

The large-scale digitisation of artworks from collections all over the world has opened the opportunity to study art from a computer vision perspective, by building tools to help in the conservation and dissemination of cultural heritage. Some of the most promising work on this direction involves the automatic analysis of paintings, in which computer vision techniques are applied to study the content \cite{crowley2016art,seguin2016visual}, the style \cite{collomosse2017sketching,sanakoyeu2018style}, or to classify the attributes \cite{mensink2014rijksmuseum,mao2017deepart} of a specific piece of art.

\begin{figure}
\vspace{15pt}
\centering
\includegraphics[width = 0.47\textwidth]{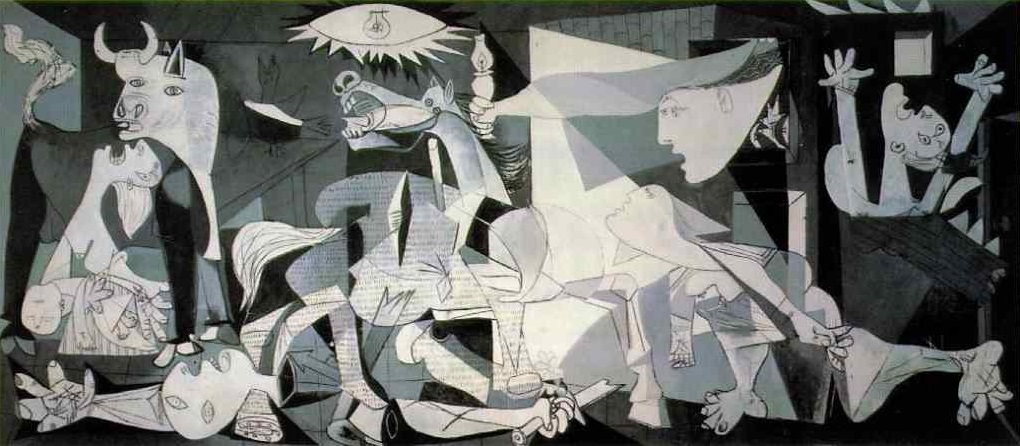}
\caption{Art as an element in a global context. In Guernica, Pablo Picasso, by means of his own style built upon many artistic influences, such as Cubism or African art, expressed his emotions against war inspired by its historical and political context. Image source: \url{www.PabloPicasso.org}.} \label{fig:guernika}
\end{figure}

Automatic analysis of art usually involves the extraction of visual features from digitised artworks by using either hand-crafted \cite{carneiro2012artistic, khan2014painting, shamir2010impressionism} or deep learning techniques \cite{karayev2014recognizing,Tan2016CeciNP,ma2017part,mao2017deepart}. Visual features, specially the ones extracted from convolutional neural networks (CNN)  \cite{krizhevsky2012imagenet,Simonyan15,he2016deep}, have been shown to be very powerful at capturing content \cite{crowley2016art} and style \cite{collomosse2017sketching} from paintings, producing outstanding results, for example, on the field of style transfer \cite{sanakoyeu2018style}. However, art specialists rarely analyse artworks as independent and isolated creations, but commonly study paintings within its artistic, historical and social context, such as the author influences or the connections between different schools, as illustrated in Figure \ref{fig:guernika}.

\begin{figure*}
\centering
\includegraphics[width = 0.8\textwidth]{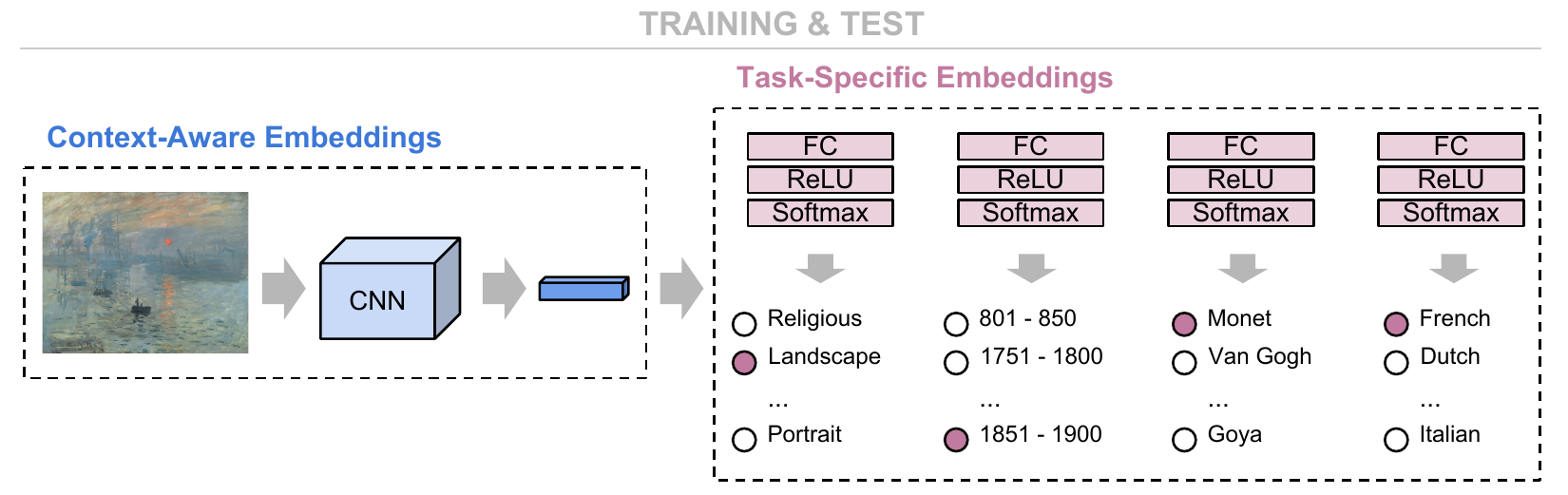}
\caption{Overview of the Multi-Task Learning model.} 
\label{fig:methods:mtmodel}
\end{figure*}

In an attempt to analyse art from a global perspective, we propose to extract context-aware embeddings from paintings by considering both visual and contextual information. We propose two independent and separate models to encode contextual data in art. In the first one, a multi-task learning model (MTL) is used to capture visual relationships between artistic attributes in paintings. To do that, we optimise several artistic-related tasks jointly, so that the obtained generic embeddings are enforced to find common elements and hence, context, between the different tasks. In the second model, in contrast, we use a knowledge graph (KG) to learn the different relationships between artistic attributes. We create an art-specific KG by connecting a set of paintings with their artistic-related attributes. Then, node neighbourhoods and positions within the graph are encoded into a vector to represent context. Whereas the MTL model is able to capture relationships occurring at the visual-level, the use of KGs offers a more flexible representation of arbitrary relationships, which might not be well-structured and more difficult to detect when considering visual content only. In any case, we incorporate the information obtained with the aforementioned models into the art analysis system.

The two proposed context-aware embeddings are evaluated on the SemArt dataset \cite{Garcia2018How} in four different art classification tasks and in two cross-modal retrieval tasks. We show that, although none of the proposed models show a superior performance with respect to the other one in all of the evaluated tasks, context-aware embeddings consistently outperform methods based on visual embeddings only. Reported results show that performance is improved by up to 7.3\% in art classification and a 37.24\% in retrieval when our methods are used. In summary, our contributions are:
\begin{itemize}
    \item To enhance standard visual features with contextual data in automatic art analysis.
    \item To propose a multi-task learning model to encode the visual relationships between different artistic attributes.
    \item To build an art-specific knowledge graph to capture contextual relationships between artistic attributes and use it to inform the visual model.
    \item To provide evidence that including contextual information improves performance in multiple art analysis tasks.
\end{itemize}

%-------------------------------------------------------------------------
\section{Related Work}
\label{sec:relatedwork}

\textbf{Automatic Art Analysis}. In order to identify specific attributes in paintings, early work in automatic art analysis was focused on representing the visual content of paintings by designing hand-crafted feature extraction methods \cite{johnson2008image,shamir2010impressionism,carneiro2012artistic,khan2014painting,mensink2014rijksmuseum}. For example, \cite{johnson2008image} proposed to detect authors by analysing their brushwork using wavelet decompositions, \cite{shamir2010impressionism, khan2014painting} combined  color, edge, or texture features for author, style, and school classification and \cite{carneiro2012artistic,mensink2014rijksmuseum} used SIFT features \cite{lowe2004distinctive} to classify paintings into different attributes.

In the last years, deep visual features extracted from CNNs have been repeatedly shown to be very effective in many computer vision tasks, including automatic art analysis \cite{Bar2014ClassificationOA,karayev2014recognizing,Saleh2015LargescaleCO,Tan2016CeciNP,ma2017part,mao2017deepart,Garcia2018How,strezoski2018omniart}. At first, deep features were extracted from pre-trained networks and used off-the-shelf for automatic art classification \cite{Bar2014ClassificationOA,karayev2014recognizing,Saleh2015LargescaleCO}. Later, deep visual features were shown to obtain better results when fine-tuned using painting images \cite{Tan2016CeciNP,seguin2016visual,mao2017deepart,strezoski2018omniart,chu2018image}. Alternatively, \cite{crowley2015face,crowley2014state,crowley2016art} explored domain transfer for object and face recognition in paintings, whereas \cite{Garcia2018How} introduced the use of joint visual and textual models to study paintings from a semantic perspective.

So far, most of the proposed methods in automatic art analysis have been focused on representing the visual essence of an artwork by capturing style and/or content. However, the study of art is not only about the visual appearance of paintings, but also about their historical, social, and artistic context. In this work, we propose to consider both visual and contextual information in art by introducing context-aware embeddings obtained by either a multi-task learning model or a knowledge graph.

\vspace{7pt}
\noindent
\textbf{Multi-Task Learning}. Multi-task learning models \cite{caruana1997multitask} aim to solve multiple tasks jointly with the hope that the generated generic features are more powerful than task-specific representations. In deep learning approaches, MTL is commonly performed via hard or soft parameter sharing \cite{ruder2017overview}. Whereas in hard parameter sharing \cite{caruana1997multitask,sener2018multi}, except by the output layers, parameters are shared between all the tasks, in soft parameter sharing \cite{long2015learning,yang2016deep}, each task is defined by its own parameters, which are encouraged to remain similar via regularisation methods. 

Following the success of MTL in many computer vision problems, such as object detection and recognition \cite{salakhutdinov2011learning,bilen2016integrated}, object tracking \cite{zhang2013robust}, facial landmark detection \cite{zhang2014facial}, or facial attribute classification \cite{rudd2016moon}, we propose a hard parameter sharing MTL approach for obtaining context-aware embeddings in the domain of art analysis. In our approach, by jointly learning related artistic tasks, the resulting visual representations are enforced to capture relationships and common elements between the different artistic attributes, such as author, school, type, or period, and thus, providing contextual information about each painting.

\vspace{7pt}
\noindent
\textbf{Knowledge Graphs}. Knowledge graphs are complex graph structures able to capture non-structured relationships between the data represented in the graph.
When KGs are used to add contextual information to a multimedia database, prior work has shown consistent improvements in annotation, classification, and retrieval benchmarks \cite{fergus2010semantic, salakhutdinov2011learning, chen2013neil, deng2014large, johnson2015image, Zhang:2016:LCI:3012406.2978656, marino2017more, cui2018general, wang2018zero}.

To extract contextual information from a KG, one strategy is to encode relationships from visual concepts detected in pictures, forming concept hierarchies \cite{salakhutdinov2011learning, deng2014large}. \cite{johnson2015image} introduced human generated scene graphs based on descriptions of pictures to improve retrieval tasks, whereas \cite{cui2018general} exploited semantic relationships between labels using ConceptNet \cite{speer2012representing}. Another strategy is to gather labelling information from social media to compute a word-image graph, in which random walks are proposed to extract topological information \cite{Zhang:2016:LCI:3012406.2978656}. Other approaches incorporate the use of external knowledge bases. For example, \cite{fergus2010semantic} proposed to improve classifiers with the use of WordNet \cite{miller1995wordnet}, \cite{marino2017more} designed an end-to-end learning pipeline to incorporate large knowledge graphs, such as Visual Genome \cite{krishnavisualgenome}, into classification, and \cite{wang2018zero} trained image and graph embeddings using WordNet, NELL \cite{carlson2010toward}, or NEIL \cite{chen2013neil}. 

While related work mostly rely on the use of external knowledge, in our knowledge graph model, we propose to capture contextual information only by processing the data provided with art datasets. As the semantic of art pieces is extremely domain specific, the symbolism that is implied in mythological or religious representations may not benefit from general knowledge. Instead, we leverage on metadata information from art datasets to create a domain-specific knowledge graph, from which we train context embeddings without any task-specific supervision.

%-------------------------------------------------------------------------
\section{Context-Aware Embeddings}
\label{sec:methods}
We propose two independent models for capturing artistic context in paintings. In the first model, 
relationships between different artistic attributes are extracted from the visual appearance of images with a multi-task learning model, whereas in the second model, non-visual information extracted from artistic metadata is encoded by using knowledge graphs.

\subsection{Multi-Task Learning Model}
\label{sec:methods:multitask}
In the MTL model, artistic context is obtained by finding visual relationships between common elements in different artistic attributes. To compute context-aware embeddings, the model is trained to learn multiple artistic tasks jointly, so the generated embeddings are enforced to find visual similarities between the different tasks.

Formally, in a multi-task learning problem, given $T$ learning tasks, with the training setting for the $t$-th task consisting on $N_t$ training samples and denoted as $\{\Vec{x}^t_j,y^t_j\}^{N_t}_{j=1}$, where  $\Vec{x}^t_j \in \mathbb{R}^d$ and $y^t_j$  are the $j$-th training sample and its label, respectively, the goal is to optimise:%
\begin{equation}
    \argmin_{\{\Vec{w}^t\}^T_{t=1}} \sum^T_{t=1} \sum^{N_t}_{j=1} \lambda^t \ell_t(f(\Vec{x}_j^t;\Vec{w}^t),y_j^t)
\end{equation} %
where $f$ is a function parameterised by the vector $\Vec{w}^t$, $\ell_t$ is the loss function for the $t$-th task, and $\lambda^t$, with $\sum^T_{t=1}\lambda_t = 1$, weights the contribution of each task.

In our model, we aim to distinguish between the context-aware information and the task-specific data. We define the function parameters for the $t$-th task as the contribution of two vectors, $\Vec{w}^t = [\vec{w}_g^t; \vec{w}_s^t]$, so that $f$ is defined as:%
\begin{equation}
    f(\Vec{x}_j^t;\Vec{w}^t) =  f_s(\vec{v}_j^t;\vec{w}_s^t)
\end{equation}%
where % 
\begin{equation}
\vec{v}_j^t = f_g(\vec{x}_j^t;\vec{w}_g^t)
\end{equation} %
with $f_g$ being a context-aware function parametrised by $\vec{w}_g^t$, $f_s$ being a task-specific function parametrised by $\vec{w}_s^t$, and $\vec{v}_j^t$ being the $j$-th context-aware embedding generated by task $t$.

By sharing both the training data and the context-aware parameters across all the tasks as $\vec{x}_j^t = \vec{x}_j^k \text{ and } \vec{w}_g^t = \vec{w}_g^k \text{ for } j \neq k$, the context-aware embedding $\vec{v}_j^t$ is defined as: %
\begin{equation}
\vec{v}_j = f_g(\vec{x}_j;\vec{w}_g)
\end{equation} %
which enforces $\vec{v}_j$ to encode $\vec{x}_j$ in a generic and non task-specific representation by identifying patterns and relationships within different tasks. The problem, finally, is formulated as: % 
\begin{equation}
    \argmin_{\vec{w}_g,\{\Vec{w}_s^t\}^T_{t=1}} \sum^T_{t=1} \sum^{N}_{j=1} \lambda^t \ell_t(f_s(f_g(\vec{x}_j;\vec{w}_g);\vec{w}_s^t),y_j^t)
\label{mtl_optimization}
\end{equation} 

\begin{figure}
\centering
\includegraphics[width = 0.4\textwidth]{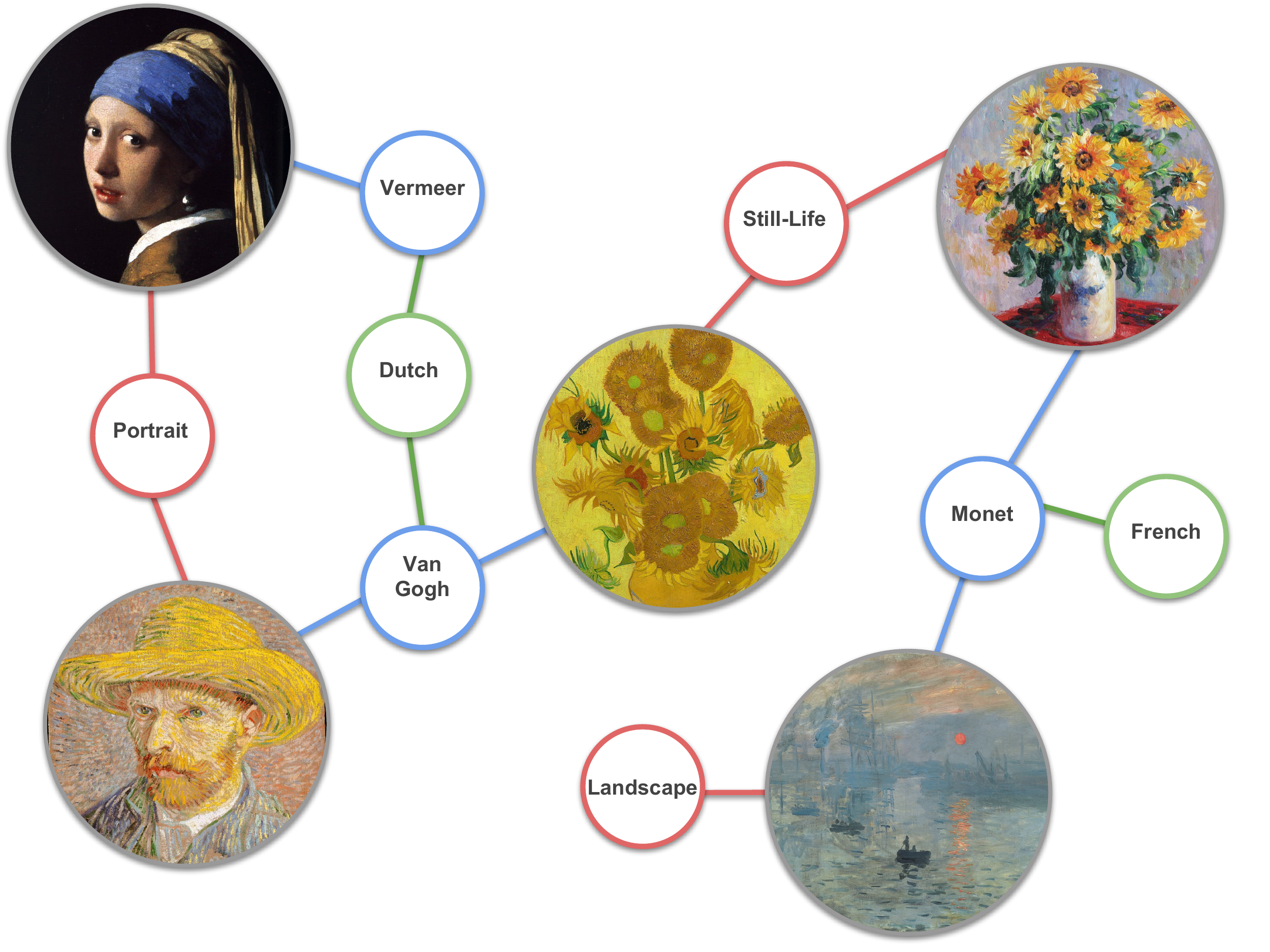}
\caption{An example of our artistic KG. Each node corresponds to either a painting or an artistic attribute, whereas edges correspond to existing interconnections.} 
\label{fig:method:exmplagraph}
\end{figure}

For solving this optimisation problem, we propose the model in Figure \ref{fig:methods:mtmodel}, in which the $T$ learning tasks correspond to multiple artistic classification challenges, such as type, school, timeframe, or author classification. To obtain context-aware embeddings, the context-aware function, $f_g$, is characterised by ResNet50 \cite{he2016deep} after removing the last fully-connected layer, whereas the task-specific functions, $f_s$, are described by a fully connected layer followed by a ReLU non-linearity. The output of $f_g$ is a 2,048 dimensional embedding, which is input into the task-specific classifiers. Each classifier produces a $C_t$-dimensional task-specific embedding as output, $\vec{z}_j^t$, where $C_t$ is the number of classes in each task. Each tasks is formulated with the cross-entropy loss function as: %
\begin{equation}
\ell_t(\vec{z}_j^t,y_j^t) = -\log \Bigg( \frac{\exp(\vec{z}_j^t[y_j^t])}{\sum_c \exp(\vec{z}_j^t[c]) } \Bigg)
\end{equation}%
where $\vec{z}_j^t = f_s(f_g(\vec{x_j};\vec{w}_g);\vec{w}_s^t)$.

\begin{figure*}
\centering
\includegraphics[width = 0.8\textwidth]{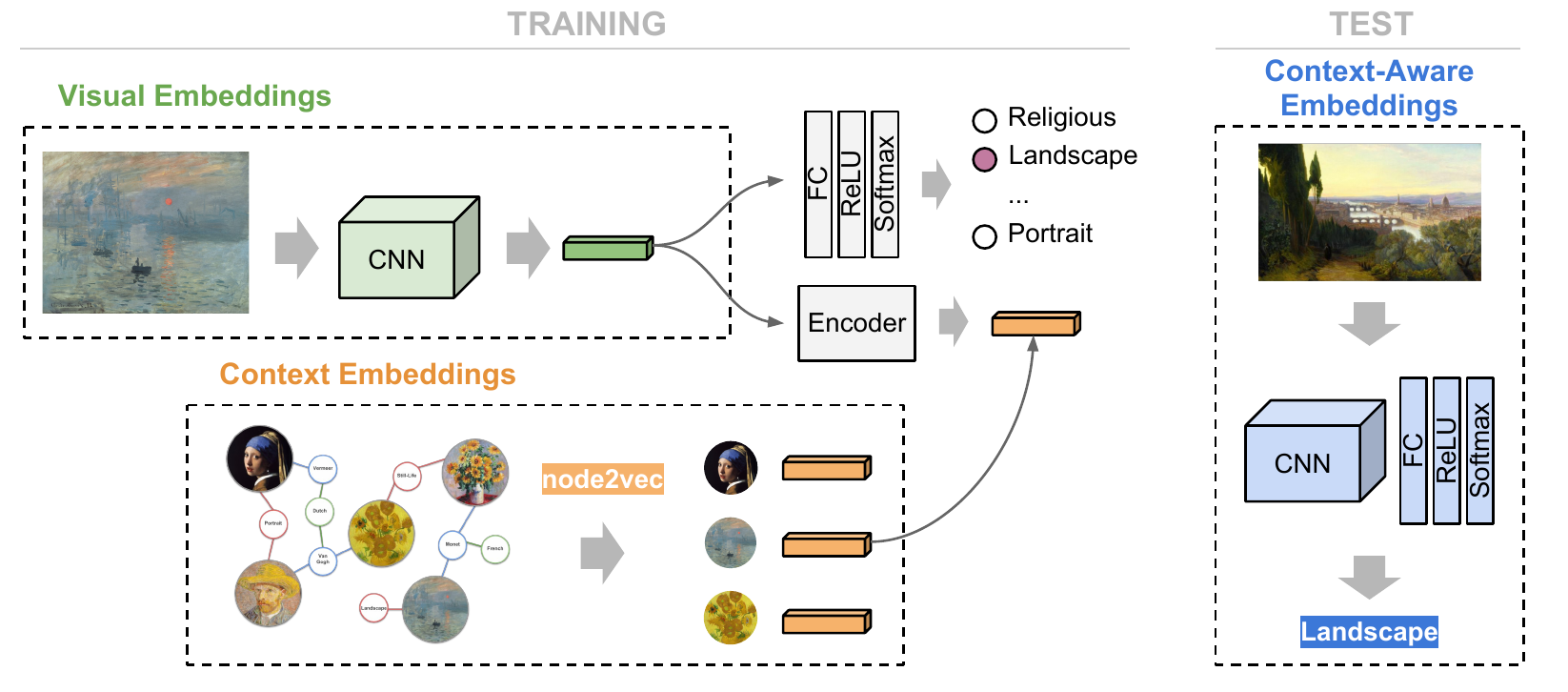}
\caption{Overview of the Knowledge Graph model.} 
\label{fig:methods:kgmodel}
\end{figure*}

\subsection{Knowledge Graph Model}
\label{sec:methods:graphs}
In the MTL model, contextual information is provided by the painting images themselves by considering the relationships between common elements in the visual appearance of the images when multiple artistic tasks are trained together. In the knowledge graph model (KGM), in contrast, contextual information is obtained from capturing relationships in an artistic knowledge graph built with non-visual artistic metadata. 

\vspace{7pt}
\noindent
\textbf{Artistic Knowledge Graph}. A KG is a graph structure, $G=(V,E)$, in which the entities and their relations are represented by a collection of nodes, $V$, and edges, $E$, respectively. We use a KG to capture contextual knowledge and similarities in the semantic space formed by the graph, often referred to as homophily \cite{goyal2018graph}. 

To construct an artistic KG, one strategy is to connect paintings with edges when sharing a common attribute $a \in A$. However, the complexity of this approach is expensive, reaching the order of $|V|^2 \times |A|$. Instead, we propose to connect paintings with their attributes in a much sparser manner. We consider multiple types of node: paintings, $P \subseteq V$, which represent the paintings themselves (e.g. Girl with a Pearl Earring), and each family, $\psi$, of attributes $A_\psi \subseteq V$, which represent artistic concepts (e.g. a type such as Portrait or an author such as Van Gogh). We use the training data from the SemArt dataset \cite{Garcia2018How}, which contains 19,244 paintings labeled with the attributes \textit{Author}, \textit{Title}, \textit{Date}, \textit{Technique}, \textit{Type}, \textit{School}, and \textit{Timeframe} to connect edges, $e=(V_p,V_q) \in E$, between painting nodes, $V_p$, and attribute nodes, $V_q \in A_\psi$, with $\psi\in \{\text{\textit{Type}, \textit{Timeframe}, \textit{Author}}\}$, when an attribute exists in a painting. As \emph{School} corresponds to an author's school, we connect an edge, $e=(V_a,V_s) \in E$, between an author, $V_a$, and a school, $V_s$. We additionally enrich our graph with three other families of attributes, which are connected to painting nodes. From \emph{Technique}, we extract \emph{Material}, such as oil, and \emph{Support}, such as 210x80cm. Also, by computing the most common $n$-grams in the titles, with $n$ up to three, we extract keywords from the title of each painting, such as Three Graces. In total, the resulting KG presents 33,148 nodes and 125,506 edges, with 3,166 authors, 618 materials, 26 schools, 8,899 supports, 22 timeframes, 10 types, and 1,163 keyword nodes. An example representation of our artistic graph is shown in Figure \ref{fig:method:exmplagraph}.

\vspace{7pt}
\noindent
\textbf{Training}. At training time, visual and context embeddings are computed from the painting image and from the KG, respectively, and used to optimise the weights of the model. Our training model is depicted in Figure \ref{fig:methods:kgmodel} and each of its parts are detailed below.

\textit{Visual Embeddings}. Visual embeddings represent the visual appearance of paintings, containing information about the content and the style of the artwork. To obtain the visual embeddings, we use a ResNet50 \cite{he2016deep} without the last fully connected layer.
    
\textit{Context Embeddings}. Context embeddings encode the artistic context of an artwork by extracting data from the KG. For encoding the KG information into a vector representation, we adopt the node2vec model \cite{grover2016node2vec} because of its capacity to preserve a trade-off between homophily and structural equivalences, resulting in high performances in node classification tasks \cite{goyal2018graph}. To capture node embeddings, node2vec operates skip-grams over random walks in the KG, and associates a vector representing the neighbourhood and the overall position of each node in the graph. 
    
\textit{Classifier}. The classifier takes as input the visual embedding and predicts the artistic attributes contained in the sample painting. We use different kinds of attribute classifiers, such as type, school, timeframe, or author. The classifier is composed of a fully connected layer followed by a ReLU non-linearity, and its output is used to compute a classification loss using a cross-entropy loss function:%
\begin{equation}
    \ell_c(\vec{z}_j,class_j) = -\log \Bigg( \frac{\exp(\vec{z}_j[class_j])}{\sum_i \exp(\vec{z}_j[i]) } \Bigg)
\label{eq:crossentropy}
\end{equation} %
where $\vec{z}_j$ and $class_j$ are the output of the classifier and the assigned label of the attribute for the $j$-th training painting, respectively.

\textit{Encoder}. The encoder module, which is composed of a single fully connected layer, is used to project the visual embeddings into the context embedding space. We compute the loss between the projected visual embedding, $\vec{p}_j$, and the context embedding, $\vec{u}_j$, of the $j$-training sample with a smooth L1 loss function: %
\begin{equation}
   \ell_e(\vec{p}_j,\vec{u}_j) = \frac{1}{n} \sum_i \delta_{ji}
\end{equation} %
where %
\begin{equation*}
    \delta_{ji} = \begin{cases}
          \frac{1}{2}(p_{ji} - u_{ji})^2, & \text{if}\ | p_{ji} - u_{ji} | \leq 1 \\
      | p_{ji} - u_{ji}  | - \frac{1}{2}, & \text{otherwise}
    \end{cases}
\end{equation*} %
being $p_{ji}$ and $u_{ji}$ the $i$-th elements in $\vec{p}_j$ and $\vec{u}_j$, respectively. To train the KGM, we compute the total loss function of the model as a combination of the losses obtained from the classifier and encoder modules: %
\begin{equation}
    \mathcal{L} = \lambda_c \sum_{j=1}^{N} \ell_c(\vec{z}_j,class_j) + \lambda_e \sum_{j=1}^{N} \ell_e(\vec{p}_j,\vec{u}_j)
    \label{eq:totalloss}
\end{equation} %
where $\lambda_c$ and $\lambda_e$ are parameters that weight the contribution of the classification and the encoder modules, respectively, and $N$ is the number of training samples. 

Whereas the parameters of the context embeddings are learnt without supervision and frozen during the KGM training process, the loss score, $\mathcal{L}$, obtained from Equation (\ref{eq:totalloss}) is backpropagated through the weights of the visual embedding module. This enforces ResNet50 to compute embeddings that are meaningful for artistic classification by decreasing $\ell_c$, while incorporating contextual information from the knowledge graph by minimizing $\ell_e$. 

\vspace{7pt}

\noindent
\textbf{Context-Aware Embeddings}. At test time, to obtain context-aware embeddings from unseen test samples, painting images are fed into the fine-tuned ResNet50 model. As context embeddings computed directly from the KG cannot be obtained for samples that are not contained as a node, the context embedding and the encoder modules are removed from the test model (Figure \ref{fig:methods:kgmodel}). 

However, the ResNet50 network has been enforced during the training process to (1) capture relevant visual information to predict artistic attributes and (2) to incorporate contextual data from the KG in the visual embeddings. Therefore, the output embeddings from the fine-tuned ResNet50 are, indeed, context-aware embeddings.

%-------------------------------------------------------------------------
\section{Art Classification}
\label{sec:classification}
We evaluated the two proposed methods in multiple art classification tasks, including author identification and type classification.

\subsection{Implementation Details}
\label{sec:class_eval:details}
In both of our proposed models, painting images are encoded into a vector representation by using ResNet50 \cite{he2016deep} without the last fully connected layer. ReNet50 is initialised with its standard pre-trained weights for image classification, whereas the weights from the rest of the layers are initialised randomly. Images are scaled down to 256 pixels per side and randomly cropped into 224 $\times$ 224 patches. At training time, visual data is augmented by randomly flipping images horizontally. The size of the embeddings produced by ResNet50 is 2,048, whereas the dimensionality produced by node2vec is 128. We use stochastic gradient descent with a momentum of 0.9 and a learning rate of 0.001 as optimiser. The training is conducted in mini-batches of 28 samples, with a patience of 30 epochs. In the MTL model, $\lambda_t$ is set to 0.25 for all the tasks, whereas in the KGM approach, $\lambda_c$ is set to 0.9 and $\lambda_c$ to 0.1.

\subsection{Evaluation Dataset}
\label{sec:class_eval:dataset}
Most of the proposed datasets for art classification evaluation are either too small to train deep learning models \cite{carneiro2012artistic,khan2014painting,crowley2014state}, do not provide multiple attributes per image \cite{karayev2014recognizing}, or are not available to download at the time of writing this work \cite{mao2017deepart}. Thus, we decide to use the recently introduced SemArt dataset \cite{Garcia2018How}, which was envisaged for cross-modal retrieval, and can be easily adapted for classification evaluation. The SemArt dataset is a collection of 21,384 painting images, from which 19,244 are used for training, 1,069 for validation and 1,069 for test. Each painting is associated with an artistic comment and the following attributes: \textit{Author}, \textit{Title}, \textit{Date}, \textit{Technique}, \textit{Type}, \textit{School} and \textit{Timeframe}. We implement the following four tasks for art classification evaluation.

\textbf{Type Classification}. Using the attribute \textit{Type}, each painting is classified according to 10 different common types of paintings: \textit{portrait}, \textit{landscape}, \textit{religious}, \textit{study}, \textit{genre}, \textit{still-life}, \textit{mythological}, \textit{interior}, \textit{historical} and \textit{other}.

\textbf{School Classification}. The \textit{School} attribute is used to assign each painting to one of the schools of art that appear at least in ten samples in the training set: \textit{Italian}, \textit{Dutch}, \textit{French}, \textit{Flemish}, \textit{German}, \textit{Spanish}, \textit{English}, \textit{Netherlandish}, \textit{Austrian}, \textit{Hungarian}, \textit{American}, \textit{Danish}, \textit{Swiss}, \textit{Russian}, \textit{Scottish}, \textit{Greek}, \textit{Catalan}, \textit{Bohemian}, \textit{Swedish}, \textit{Irish}, \textit{Norwegian}, \textit{Polish} and \textit{Other}. Paintings with a school different to those are assigned to the class \textit{Unknown}. In total, there are 25 school classes.

\textbf{Timeframe Classification}. The attribute \textit{Timeframe}, which corresponds to periods of 50 years evenly distributed between 801 and 1900, is used to classify each painting according to its creation date. We only consider timeframes with at least ten paintings in the training set, obtaining a total of 18 classes, which includes an \textit{Unknown} class for timeframes out of the selection.

\textbf{Author Identification}. The \textit{Author} attribute is used to classify paintings according to 350 different painters. Although the SemArt dataset provides 3,281 unique authors, we only consider the ones with at least ten paintings in the training set, including an \textit{Unknown} class for painters not contained in the final selection.

\subsection{Baselines}
\label{sec:class_eval:baselines}
Our models are compared against the following baselines:

\textbf{Pre-trained Networks}. VGG16 \cite{Simonyan15}, ResNet50 \cite{he2016deep} and Res-Net152 \cite{he2016deep} with their pre-trained weights learnt in natural image classification. To adapt the models for art classification, we modified the last fully connected layer to match the number of classes of each task. The weights of the last layer were initialised randomly and fine-tuned during training, whereas the weights of the rest of the network were frozen.
    
\textbf{Fine-tuned Networks}. VGG16 \cite{Simonyan15}, ResNet50 \cite{he2016deep} and Res-Net152 \cite{he2016deep} networks were fine-tuned for each art classification task. As in the pre-trained models, the last layer was modified to match the number of classes in each task.
    
\textbf{ResNet50+Attributes}. The output of each fine-tuned classification model from above was concatenated to the output of a pre-trained ResNet50 network without the last fully connected layer. The result was a high-dimensional embedding representing the visual content of the image and its attribute predictions. The high-dimensional embedding was input into a last fully connected layer with ReLU to predict the attribute of interest. Only the weights from the pre-trained ResNet50 and the last layer were fine-tuned, whereas the weights of the attribute classifiers were frozen.
    
\textbf{ResNet50+Captions}. For each painting, we generated a caption using the captioning model from \cite{xu2015show}. Captions were represented by a multi-hot vector with a vocabulary size of 5,000 and encoded into a 512-dimensional embedding with a fully connected layer followed by an hyperbolic tangent or tanh activation. The caption embeddings were then concatenated to the output of a ResNet50 network without the last fully connected layer. The concatenated vector was fed into a fully connected layer with ReLU to obtain the prediction.

\begin{table}
\renewcommand{\arraystretch}{1.2}
\setlength{\tabcolsep}{7pt}
\caption{Art Classification Results on SemArt Dataset.}
\centering
\begin{tabular}{ l c c c c}
\hline
\textbf{Method}  & \textbf{Type} & \textbf{School} & \textbf{TF} & \textbf{Author} \\ 
\hline

VGG16 \footnotesize{pre-trained} &	0.706 & 0.502 & 0.418 & 0.482 \\
ResNet50 \footnotesize{pre-trained} &	0.726 & 0.557 & 0.456 & 0.500 \\
ResNet152 \footnotesize{pre-trained} &	0.740 & 0.540 & 0.454 & 0.489 \\

VGG16 \footnotesize{fine-tuned} & 0.768 &	0.616 &	0.559 &	0.520 \\
ResNet50 \footnotesize{fine-tuned} & 0.765 &	0.655 &	0.604 &	0.515 \\
ResNet152 \footnotesize{fine-tuned} & 0.790 &	0.653 &	0.598 &	0.573 \\

\small ResNet50+Attributes & 0.785 &	0.667 &	0.599 &	0.561 \\
\small ResNet50+Captions & 0.799 &	0.649 &	0.598 &	0.607 \\

MTL \footnotesize{context-aware} &	0.791 &	\textbf{0.691} & \textbf{0.632} & 0.603 \\
KGM \footnotesize{context-aware}  &	\textbf{0.815} &	0.671 & 0.613 & \textbf{0.615} \\
\hline

% \Xhline{2\arrayrulewidth}
\end{tabular}
\label{tab:classification}
\end{table}

\subsection{Results Analysis}
\label{sec:class_eval:results}
We measured classification performance in terms of accuracy, i.e. the ratio of correctly classified samples over the total number of samples. Results are provided in Table \ref{tab:classification}. In every task, the best accuracy was obtained when a context-aware embedding method, MTL or KGM, was used. The MTL model performed slightly better than the KGM in \emph{School} and \emph{Timeframe} tasks, whereas the KGM was the best in classifying \emph{Type} and \emph{Author} attributes. Unsurprisingly, the pre-trained models obtained the worst results among all the baselines, as they do not present enough discriminative power in the domain of art. Also, there was a clear improvement with respect to pre-trained baselines when the networks were fine-tuned, as already noted in previous work  \cite{Tan2016CeciNP,seguin2016visual,mao2017deepart,strezoski2018omniart}. On the other hand, adding attributes or captions to the visual representations seemed to improve the accuracy, although not in all the scenarios, e.g., \emph{Timeframe} was better classified with the fine-tuned ResNet50 model than with ResNet50+Attributes or ResNet50+Captions, whereas \emph{School} was better classified with the fine-tuned ResNet50 than with ResNet50+Captions. This suggests that informing the model with extra information may be beneficial. When the data used to inform the model was contextual and obtained from capturing relationships in artistic attributes, accuracy was boosted, with improvements ranging from 3.16\% to 7.3\% with respect to  fine-tunned networks and from 1.32\% to 5.5\% with respect to ResNet50+Attributes and ResNet50+Captions models.

\section{Art Retrieval}
\label{sec:retrieval}
We also evaluated the use of context-aware embeddings on art retrieval problems by incorporating context-aware embeddings into a cross-modal retrieval algorithm. 

\subsection{Implementation Details}
\label{sec:retrieval:details}
As evaluation protocol, we used the SemArt dataset and its proposed Text2Art challenge, which consists of two cross-modal retrieval tasks: text-to-image and image-to-text. In text-to-image retrieval, given an artistic comment and its attributes, the goal is to find the correct painting within all the test paintings in the dataset. Similarly, in image-to-text retrieval, given a sample painting, the goal is to find the correct comment. Details on the dataset attributes and data splits are described in Section \ref{sec:class_eval:dataset}. We incorporate context-aware embeddings in a cross-modal retrieval model by adding the context-aware classifiers in the visual encoder module. Details of the cross-modal retrieval model are provided below.

\textbf{Visual encoder}: Painting images are scaled down to 256 pixels per side and randomly cropped into 224 $\times$ 224 patches. Then, paintings are fed into ResNet50, initialised with its standard pre-trained weights, to obtain a 1,000-dimensional vector, $\vec{h}_{\text{cnn}}$, from the last convolutional layer. At the same time, paintings are fed into a context-aware classifier (i.e. the MTL model from Figure \ref{fig:methods:mtmodel} or the KGM model from Figure \ref{fig:methods:kgmodel}) to obtain a $c$-dimensional vector, $\vec{h}_{\text{att}}$, containing the predicted attributes, with $c$ being the number of output classes in the classifier. The final visual representation, $\vec{h}$, is then computed as $\vec{h} = \vec{h}_{\text{cnn}} \oplus \vec{h}_{\text{att}}$, where $\oplus$ is concatenation.
    
\textbf{Comment and attribute encoder}: We encode each comment as a term frequency - inverse document frequency (tf-idf) vector, $\vec{q}_{\text{com}}$, using a vocabulary of size 9,708, which is built with the alphabetic words that appear at least ten times in the training set. We encode titles as another tf-idf vector, $\vec{q}_{\text{tit}}$, with a vocabulary of size 9,092, which is built with the alphabetic words that appear in the titles of the training set. Additionally, we encode \textit{Type}, \textit{School}, \textit{Timeframe}, or \textit{Author} attributes using a $c$-dimensional one-hot vector, $\vec{q}_{\text{att}}$, with $c$ being the number of classes in each attribute. The final joint comment and attributes representation, $\vec{q}$, is computed as $\vec{q} = \vec{q}_{\text{com}} \oplus \vec{q}_{\text{tit}} \oplus \vec{q}_{\text{att}}$.
    
\textbf{Cross-Modal Projections}: To compute similarities between cross-modal data, the visual representation, $\vec{h}$, and the joint comment and attributes representation, $\vec{q}$, are projected into a common 128-dimensional space using the non-linear functions $f_h$ and $f_q$, respectively. The non-linear functions are implemented with a fully connected layer followed by tanh activation and a $\ell_2$-normalisation. Once projected into the common space, elements are retrieved according to their cosine similarity.

\begin{table}
\renewcommand{\arraystretch}{1.2}
\setlength{\tabcolsep}{2pt}
\caption{Results on the Text2Art Challenge.}
\centering
\begin{tabular}{ l c c c c c c c c c c}
% @{\hspace{1em}}c@{\hspace{1em}}

\hline 
 & \multicolumn{4}{c}{\textbf{Text-to-Image}} & & \multicolumn{4}{c}{\textbf{Image-to-Text}} \\ \cline{2-5} \cline{7-10}

\textbf{Model} & \textbf{R@1} & \textbf{R@5} & \textbf{R@10} & \textbf{MR} & & \textbf{R@1} & \textbf{R@5} & \textbf{R@10} & \textbf{MR} \\ \hline 

CML & 0.144 & 0.332 & 0.454 & 14 & & 0.138 & 0.327 & 0.457 & 14 \\
CML* & 0.164 & 0.384 & 0.505 & 10 & & 0.162 & 0.366 & 0.479 & 12 \\
\hline
AMD \scriptsize{Type} & 0.114 & 0.304 & 0.398 & 17 & & 0.125 & 0.280 & 0.398 & 16  \\
AMD \scriptsize{School} & 0.103 & 0.283 & 0.401 & 19 & & 0.118 & 0.298 & 0.423 & 16 \\
AMD \scriptsize{TF} & 0.117 & 0.297 & 0.389 & 20 & & 0.123 & 0.298 & 0.413 & 17 \\
AMD \scriptsize{Author} & 0.131 & 0.303 & 0.418 & 17 & & 0.120 & 0.302 & 0.428 & 16 \\
\hline
Res152 \scriptsize{Type} & 0.178 & 0.383 &	0.525 &	9 & & 0.165 &	0.364 &	0.491 &	11 \\
Res152 \scriptsize{School} & 0.192 & 0.386 & 0.507 & 10 & & 0.163 & 0.364 & 0.484 & 12 \\ 
Res152 \scriptsize{TF} & 0.127 & 0.322 & 0.432 & 18 & &	0.130 & 0.336 &	0.444 &	16 \\
Res152 \scriptsize{Author} & 0.236	& 0.451 & 0.572 & 7 & &	0.204 & 0.440 & 0.535 &	8 \\
\hline
MTL \scriptsize{Type} & 0.145	& 0.358 & 0.474 & 12 & & 0.150 & 0.350 & 0.475 & 12\\
MTL \scriptsize{School} & 0.196 & 0.428 &	0.536 &	8 & & 0.172 &	0.396 &	0.520 &	10\\
MTL \scriptsize{TF} & 0.171 & 0.394 & 0.525 & 9 & & 0.138 & 0.353 & 0.466 & 12\\
MTL \scriptsize{Author} & 0.232 & 0.452 &	0.567 & 7 & & 0.206 & 0.431 & 0.535 & 9\\
\hline
KGM \scriptsize{Type} & 0.152 & 0.367 & 0.506 & 10 & & 0.147 & 0.367 & 0.507 & 10\\
KGM \scriptsize{School} & 0.162 & 0.371 & 0.483 & 12 & & 0.156	& 0.355 & 0.483 & 11\\
KGM \scriptsize{TF} & 0.175 & 0.399 & 0.506 & 10 & & 0.148 & 0.360 & 0.472 & 12 \\
KGM \scriptsize{Author} & \textbf{0.247} & \textbf{0.477} & \textbf{0.581} &	\textbf{6} & & \textbf{0.212} &	\textbf{0.446} &	\textbf{0.563} & \textbf{7}\\
\hline
\end{tabular}
\label{tab:retrieval_results}
\end{table}

The weights of the retrieval model, except from the context-aware classifier which is frozen, are trained using both positive (i.e. matching) and negative (i.e. non-matching) pairs of samples with the cosine margin loss function: %
\begin{equation}
\begin{split}\mathcal{L}(\vec{h}_k, \vec{q}_j) =
\begin{cases}
1 - \text{sim}(f_h(\vec{h}_k), f_q(\vec{q}_j)), & \text{if } k = j \\
\max(0, \text{sim}(f_h(\vec{h}_k), f_q(\vec{q}_j)) - \Delta), & \text{if } k \neq j
\end{cases}\end{split}
\end{equation} where $\text{sim}$ is the cosine similarity between two vectors and $\Delta = 0.1$ is the margin. We use Adam optimiser with learning rate 0.0001.

\subsection{Results Analysis}
\label{sec:retrieval:results}
Results are reported as median rank (MR) and recall rate at $K$ (R@K), with $K$ being 1, 5, and 10. MR is the value separating the higher half of the relevant ranking position amount all samples, i.e. the lower the better, whereas R@K is the rate of samples for which its relevant image is in the top $K$ positions of the ranking, i.e. the higher the better. 

We report results of the proposed cross-modal retrieval model using the following context-aware classifiers: MTL-Type, MTL-Timframe, MTL-School, MTL-Author, KGM-Type, KGM-School, KGM-Timeframe, and KGM-Author, in which only the specified attribute is used. As a baseline of the proposed model, results when using fine-tuned ResNet152 instead of a context-aware classifier are also reported. Our methods are compared against previous work: CML \cite{Garcia2018How}, which encodes comments and titles without attribute information,  
and AMD \cite{Garcia2018How}, in which attributes are used at training time to learn the visual and textual projections. CML*, is a re-implementation of CML with slightly better results. 

Results are summarised in Table \ref{tab:retrieval_results}. The KGM-Author model obtained the best results, improving previous state of the art, CML*, by a 37.24\% in average. When comparing context-aware models, in agreement with classification results (Table \ref{tab:classification}), MTL performed better than KGM when using \textit{School} whereas KGM was the best in \textit{Type} and \textit{Author} attributes. We also noted that concatenating the output of an attribute classifier as proposed (ResNet152, MTL, and KGM models) improved results considerably with respect to AMD. However, we observed a big difference in performance when using the different attributes, being \textit{Author} and \textit{Type} the best and the worst ones, respectively. A possible explanation for this phenomenon may lay in the difference on the number of classes of each attribute.

Finally, our best model, KGM-Author, was further compared against human evaluators. In the easy setup, evaluators were shown an artistic comment, a title, and the attributes \textit{Author}, \textit{Type}, \textit{School}, and \textit{Timeframe} and were asked to choose the most appropriate painting from a pool of ten random images. In the difficult setup, however, instead of random paintings, the images shown shared the same attribute \textit{Type}. Results are provided in Table \ref{tab:human}. Our model reached values closer to human accuracy than previous work, outperforming CML by a 10.67\% in the easy task and a 9.67\% in the difficult task.

\begin{table}
\renewcommand{\arraystretch}{1.3}
\setlength{\tabcolsep}{4pt}
\caption{Comparison against human evaluation on Text2Art.}
\centering

\begin{tabular}{ c l | c c c c c | c }
\hline
& \textbf{Model} & \textbf{Land} & \textbf{Relig} & \textbf{Myth} & \textbf{Genre} & \textbf{Port} & \textbf{Total} \\ 
\hline

\parbox[t]{8pt}{\multirow{4}{*}{\rotatebox[origin=c]{90}{Easy Set}}} 
& CCA \cite{Garcia2018How} & 0.708 & 0.609 & 0.571 & 0.714 & 0.615 & 0.650\\ 
& CML \cite{Garcia2018How} & \textbf{0.917} & 0.683 & 0.714 & \textbf{1} & 0.538 & 0.750 \\ 
& KGM \scriptsize{Author} & 0.875 & \textbf{0.805} & \textbf{0.857} & 0.857 & \textbf{0.846} & \textbf{0.830} \\
% \hdashline
\cline{2-8}
& Human & 0.918 & 0.795 & 0.864 & 1 & 1 & 0.889\\
\hline

\parbox[t]{8pt}{\multirow{4}{*}{\rotatebox[origin=c]{90}{Difficult Set}}} 
& CCA \cite{Garcia2018How} & \textbf{0.600} & 0.525 & 0.400 & 0.300 & 0.400 & 0.470  \\ 
& CML \cite{Garcia2018How} & 0.500 & \textbf{0.875} & 0.600 & 0.200 & 0.500 & 0.620  \\ 
& KGM \scriptsize{Author} & \textbf{0.600} & 0.825 & \textbf{0.700} &  \textbf{0.400} & \textbf{0.650} & \textbf{0.680} \\
% \hdashline
\cline{2-8}
& Human & 0.579 & 0.744 & 0.714 & 0.720 & 0.674 & 0.714 \\
\hline

\end{tabular}
\label{tab:human}
\end{table}
\section{Discussion}
\label{sec:discussion}

\begin{figure*}[htbp]
%\vspace{-5pt}
\centering
\includegraphics[width=\textwidth]{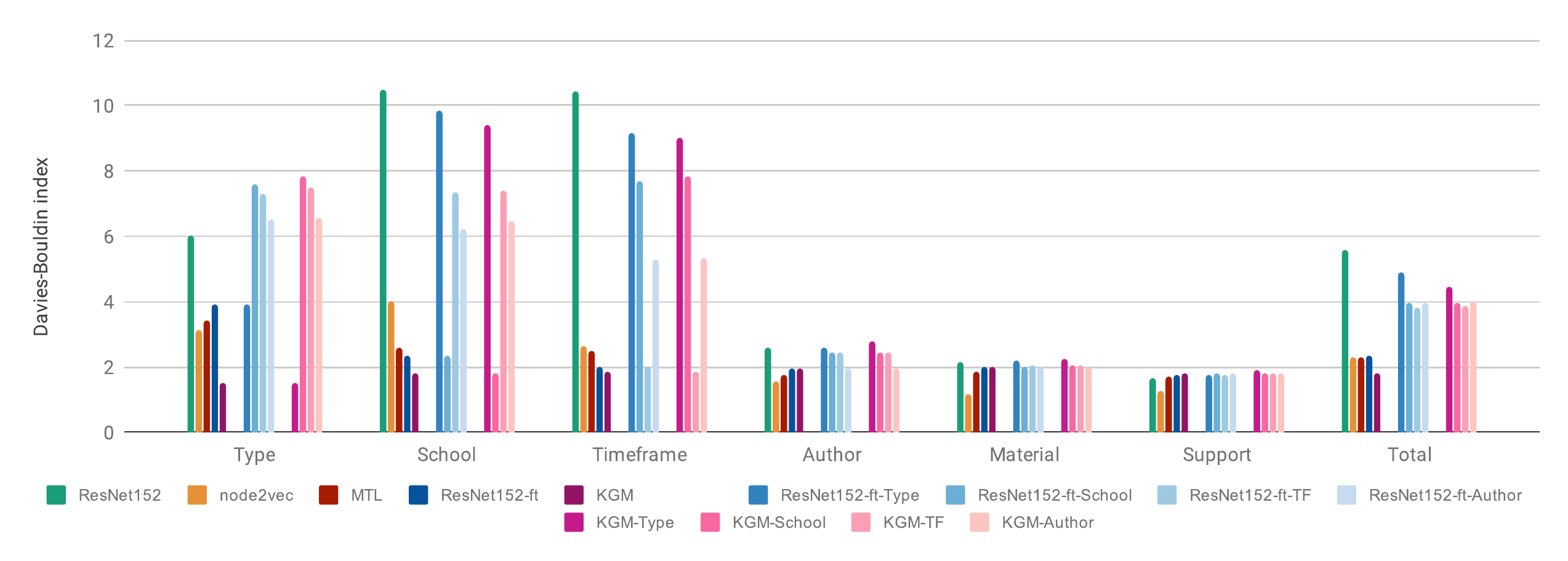}
\vspace{-15pt}
\caption{Davies-Bouldin index for each different attribute. The blue and red groups corresponds to single task of ResNet152-ft and KGM respectively. Their best results are reported in both ResNet152-ft and KGM columns of the first group.} 
\label{fig:discussion:davies}
\end{figure*}

\begin{figure*}
\centering
\vspace{20pt}
\setlength{\tabcolsep}{15pt}
\begin{tabular}{ccc}
\includegraphics[width=0.28\textwidth]{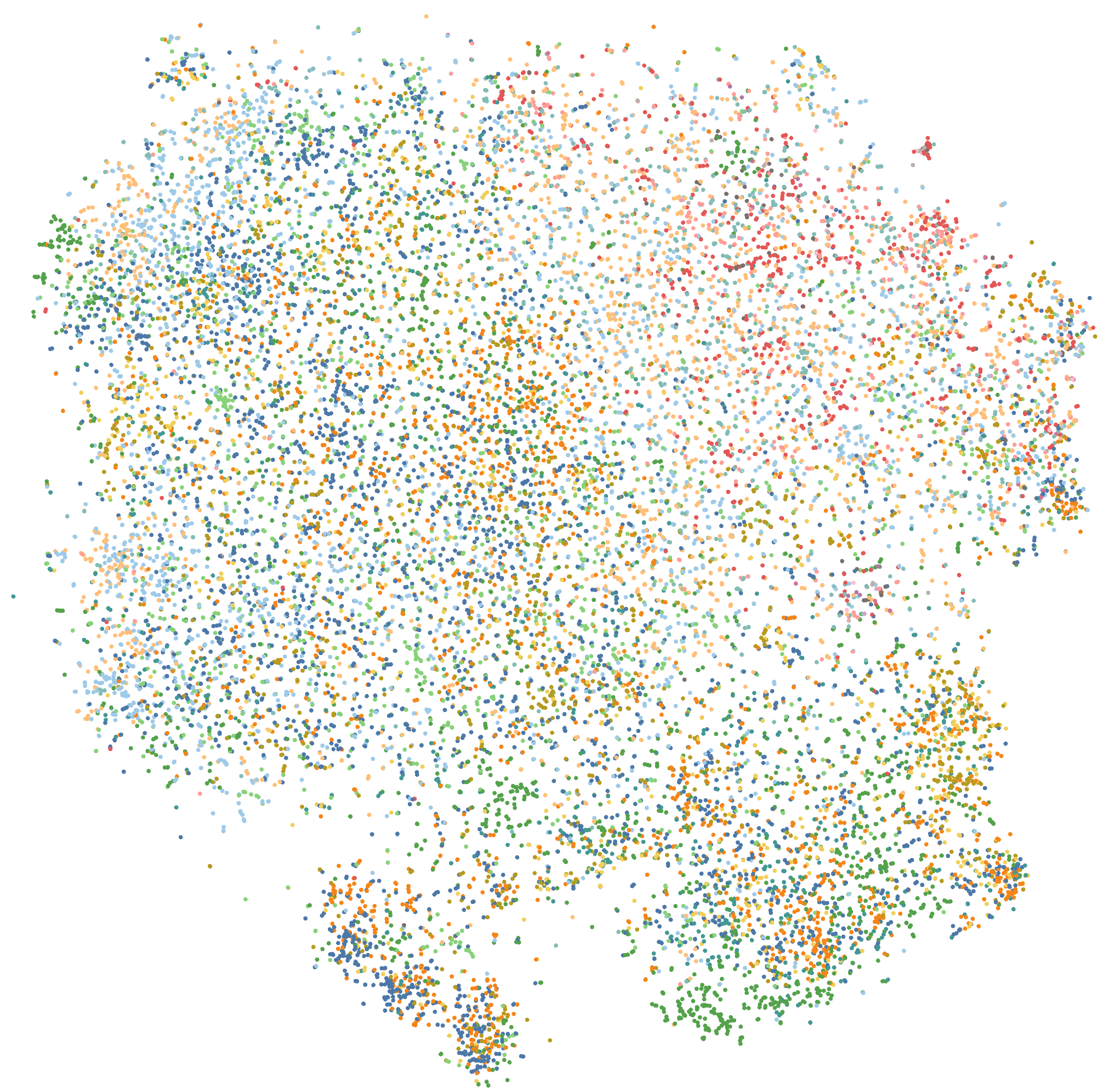} &
\includegraphics[width=0.29\textwidth]{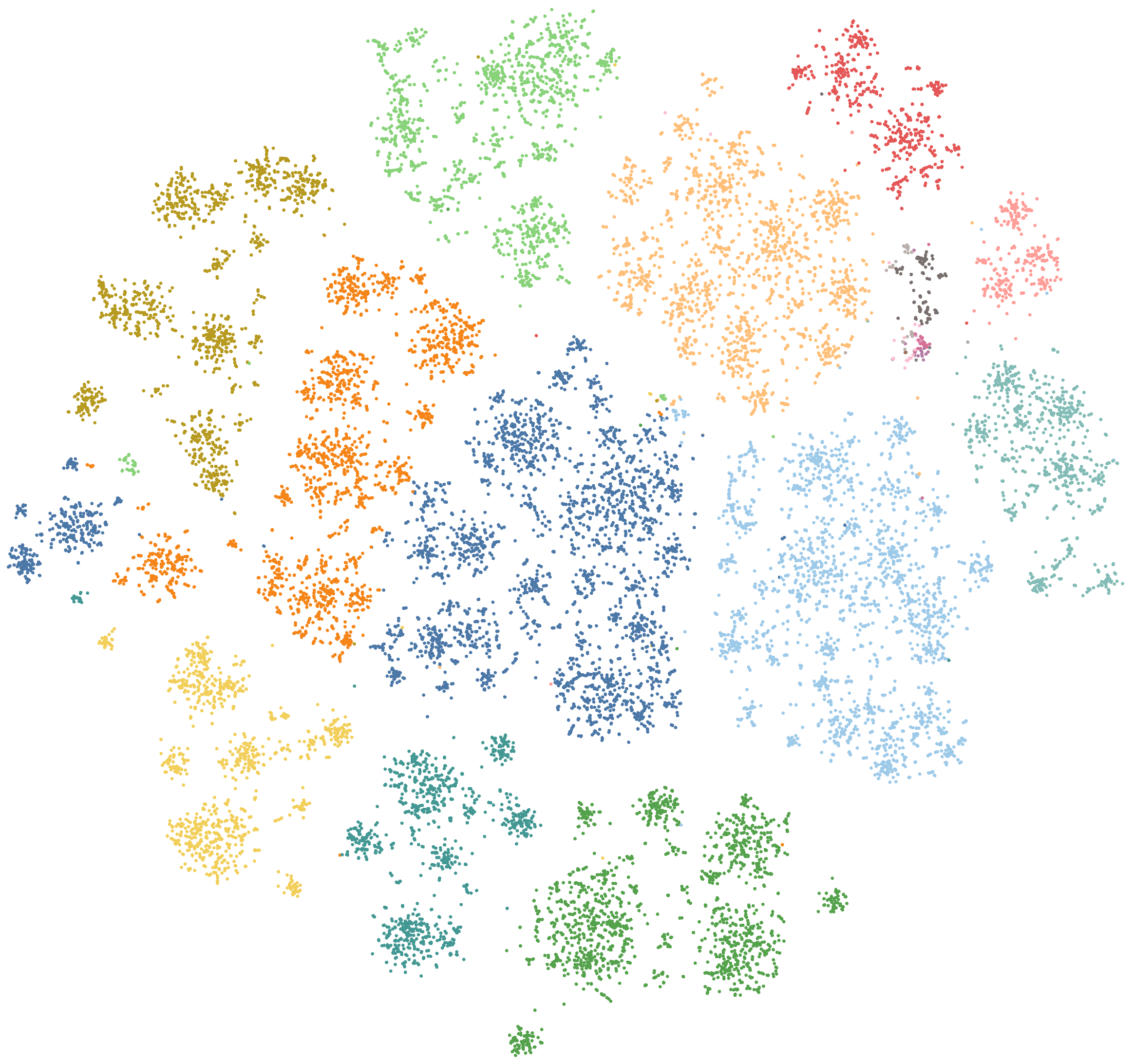} &
\includegraphics[width=0.27\textwidth]{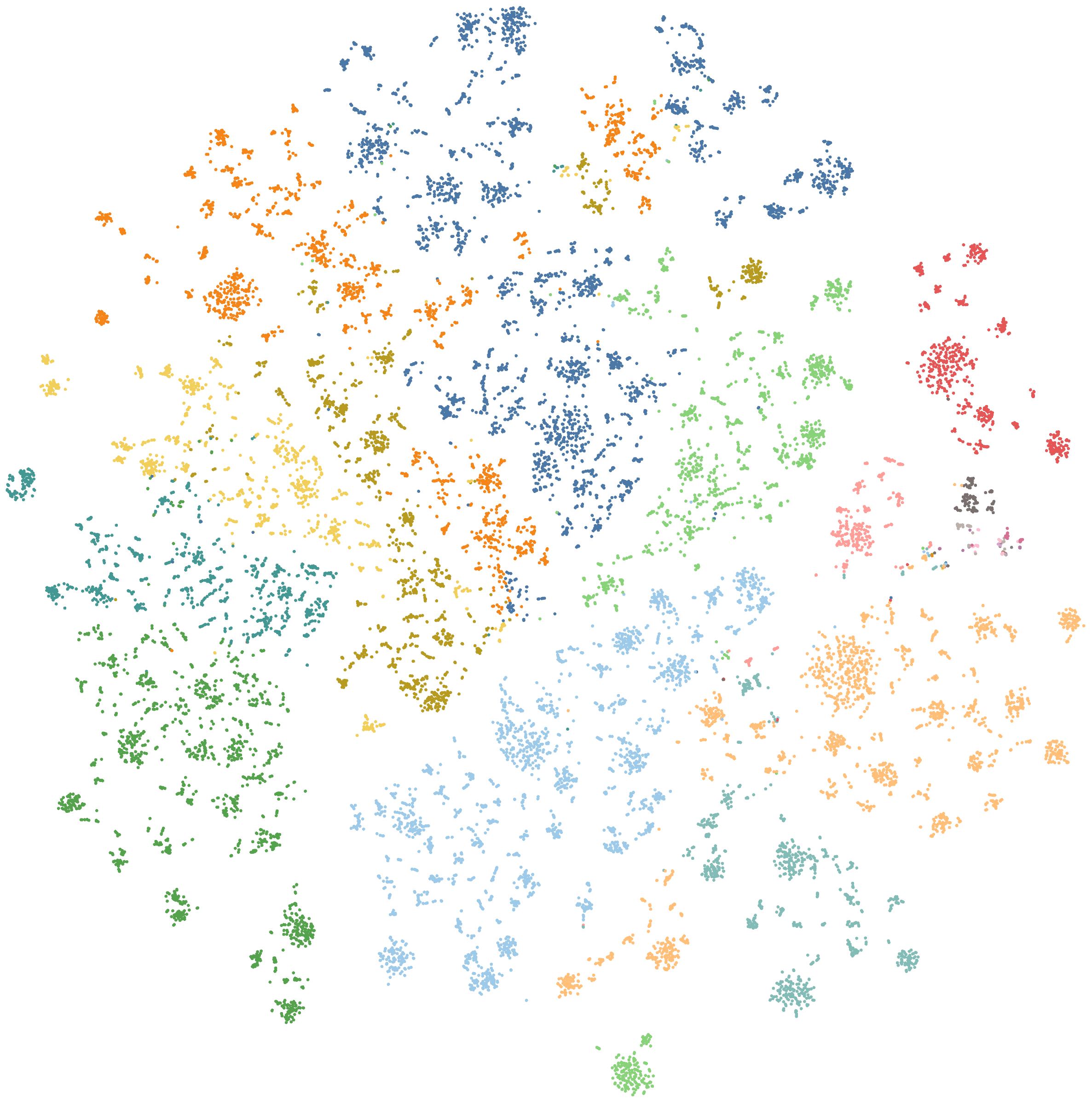} \\ [3pt]
(a) ResNet152 pre-trained &
(b) node2vec &
(c) MTL \\
\end{tabular}
\caption{Embeddings of paintings projected in Tulip \cite{auber2017tulip} using $t-$SNE \cite{wattenberg2016how}. Each node is a painting and the colouring is mapped to the \emph{Timeframe} attribute. There is good separability of \emph{Timeframe} values in the node2vec and MTL, as opposed to ResNet152.}
\label{fig:discussion:embeddings}
\end{figure*}

So far, we have presented different models to capture contextual information in art, reporting results in art classification and retrieval tasks. However, further discussion about the quality of the contextual embeddings is required. In this section, we study how well context is captured in different types of embeddings by analysing the separability of artistic attributes in clusters. 

To estimate the separability between clusters, we applied the Davies-Bouldin index \cite{davies1979cluster}, $Q$, which measures a trade-off between dispersion, $S_i$, and separation, $D_{ij}$, of the clusters $i$ and $j$: 
\begin{equation}
Q=\frac{1}{k}\sum_{i=1}^{k}\left( \max_{i \neq j} \left( \frac{S_i+S_j}{D_{ij}} \right) \right)
\end{equation}
where $k$ is the number clusters, and $S_i$ and $D_{ij}$ are computed as: 
\begin{equation*}
    S_i = \left(\frac{1}{|C_i|} \sum_{\vec{x} \in C_i}^{}{\|\vec{x} - A_i\|^p}\right)^{1/p} \qquad
    D_{ij} = \|A_i-A_j\|_p
\end{equation*}
being $A_i$ the centroid of cluster $i$ of element $\vec{x} \in C_i$ computed using the $\ell_p$ distance, and $|C_i|$ the number of elements in $C_i$.

To compare the different settings, we used the samples from the training set and we applied $Q$ with $p = 2$ to multiple types of embeddings on different attributes, as reported in Figure \ref{fig:discussion:davies}. When compared on the same task, the smaller value of $Q$, the better the cluster separation tends to be. We used \textit{Type}, \textit{School}, \textit{Timeframe}, and \textit{Author} attributes to compare performances between models. We also included the derived \emph{Material} and \emph{Support} attributes, for which none of our models was fine-tuned. Along with \emph{Author}, these new attributes have the highest dispersion due to their large number of classes, showing the lowest $Q$ values.

%Although absolute value are not of interest for us, we compare between embedding/clustering settings such as a smaller value of $Q$ means that the embedding tends to better separate clusters.  We use each painting attribute to compare performances between models w.r.t this attribute. We also include the derived \emph{Material} and \emph{Support} attributes, for which none of our models has been fine-tuned. Along with \emph{Author}, these new attributes have the highest dispersion due to their large number of classes, showing the lowest $Q$ values. 

The compared embeddings are detailed in Figure \ref{fig:discussion:davies}. The pre-trained ResNet152 baseline (in green) shows consistently the worst results in most categories, whereas the node2vec baseline trained on our KG (in orange) shows a good trade-off between categories and the best performance on the most complex attributes \emph{Author}, \emph{Material} and \emph{Support}. On average, KGM (in purple) performs the best due to its high quality on each of the \emph{Type},  \emph{School}, and \emph{Timeframe} attributes for which it has been trained. On average, the MTL (in red) shows a comparable performance to the multiple single-task fine-tuned ResNet152 (in blue).

% The pre-trained ResNet152 (in green) is our baseline, and shows consistent worse results in most categories. Node2vec trained on our KG makes a second baseline showing a nice trade-off between categories and the best performances on the most complex attributes \emph{Author}, \emph{Material} and \emph{Support}.
% On average, KGM performs best due to its high performances on each of the \emph{Type},  \emph{School}, and \emph{Timeframe} attributes for which it has been trained. On average, the MTL displays performances comparable to the multiple single-task fine-tuned ResNet152.

These results rule in favour of the added value that contextual knowledge brought by the KG improves overall performances. We may further confirm this intuition from the 2D-projected embeddings in Figure \ref{fig:discussion:embeddings}: while the space represented by pre-trained ResNet152 applied to art does not show any convincing separability, the subspace formed by paintings in the node2vec embeddings shows clear separability and sub-densities. MTL does display such a structure, while being much more fractioned. Although intriguing, the analysis of these clusters' content and how they may capture abstract concepts of art is a discussion we leave for future work.

%-------------------------------------------------------------------------
\section{Conclusions}
\label{sec:conclusions}
This work proposed the use of context-aware embeddings for art classification and retrieval problems. Two methods for obtaining context-aware embeddings were introduced. The first one, based on multi-task learning, captures the relationships between visual artistic elements in paintings, whereas the second one, based on knowledge graphs, encodes the interconnections between non-visual artistic attributes. The reported results showed that context-aware embeddings are beneficial in many automatic art analysis problems, improving art classification accuracy by up to a 7.3\% with respect to classification baselines. In cross-modal retrieval tasks, our best model outperforms previous work by a 37.24\%. Future work will study the combination of multi-task learning and knowledge graphs into a single context-aware framework.

\begin{acks}
This work was supported by JSPS KAKENHI Grant No.~18H03264.
\end{acks}

% The next two lines define the bibliography style to be used, and the bibliography file.
\bibliographystyle{ACM-Reference-Format}
\bibliography{bibliography}

\end{document}